# Highlight Timestamp Detection Model for Comedy Videos via Multimodal Sentiment Analysis

Huang Fan, WKWSCI, NTU, Singapore, fhuang004@e.ntu.edu.sg

*Abstract*— Nowadays, the videos on the Internet are prevailing. The precise and in-depth understanding of the videos is a difficult but valuable problem for both platforms and researchers. The existing video understand models do well in object recognition tasks but currently still cannot understand the abstract and contextual features like highlight humor frames in comedy videos. The current industrial works are also mainly focused on the basic category classification task based on the appearances of objects. The feature detection methods for the abstract category remains blank. A data structure that includes the information of video frames, audio spectrum and texts provide a new direction to explore. The multimodal models are proposed to make this in-depth video understanding mission possible. In this paper, we analyze the difficulties in abstract understanding of videos and propose a multimodal structure to obtain state-of-the-art performance in this field. Then we select several benchmarks for multimodal video understanding and apply the most suitable model to find the best performance. At last, we evaluate the overall spotlights and drawbacks of the models and methods in this paper and point out the possible directions for further improvements.

## I. Introduction

Video records and reflects many aspects of our social and personal lives. With the help of prevailing video websites and software, people are making and absorbing more and more new videos every day. In one single day, people watch about 1 billion hours of videos on YouTube and generate billions of comments currently. The world is producing and consuming a massive amount of video contents. And the search engine in the video website are helping people to find their interested videos in a more efficient and accurate manner. However, in most of the provided search services, video retrieval and ranking are performed by matching query terms with metadata and other video level labels. There might be several highlights or themes in one video but without obvious specifications or labels. Users may miss short but essential moments because of the lack of reminders. The supplemental timestamps information will enable many improvements for both user experience and business interests of the platform.

In recent years, people have been increasingly facing stress and depression [1]. Depression may cause many negative consequences that are harmful to both mental and physical health. The positive emotions and feelings could make people obtain a better mental and physical condition. People do need the comedy contents to erase the possible negative emotions and tend to need more comedy elements to improve their mental and physical health. The positive feedback from watching comedy videos would then push the users to absorb more, which correlates with the growing production and consumption of comedy videos in fast-growing online video-based platforms. The comedy video uploaders could also get a huge quantity of subscribers increase within only few days like Uncle Roger. The better detection and understanding of the features within comedy videos could thus help the platforms to provide better recommendation and search results for users. A better depicted user portrait then contributes to improve user stickiness and advertisement effect, which comes with more business interests and market share.

Researchers and competition participants have contributed significantly to developing video-level annotation using unconstrained and constrained models. In 2019, researchers localized video-level labels to the precise time where the label appears using a 1.53TB dataset (known as Youtube-8M dataset). However, those implementations make classifications in a more general and appearance-related level. The categories are divided by physical features like bus, ball, car and so on. The model would help the platform to better label the videos' timestamps by their physical features and categories but cannot understand the abstract and contextual meanings of specific themes.

The newly proposed multimodal structure method is a possible good solution to the abstract pattern recognition tasks for its combined multi-dimension network that can process the video, audio and text information at the same time. For one part, the text information could greatly help the model to understand the sentiment polarities and contextual features through the help of pre-trained BERT model. For another part, the audio spectrum could be a wonderful supplementation information for the video features. The exaggerated features will help the model to receive the comedy features in a more accurate way.

## II. Related Works

### A. Traditional video understanding models

The existing works of video understanding currently mainly focus on object detection using convolution neural network and pattern recognition using recurrent neural network. Those works build a foundation of general recognition of video features but still cannot understand the abstract information. The movie without audio can express the comedy features and make audience laugh. Even if the describing texts of silent movies are removed, people still can capture the humor elements. The humor elements can be recognized through either exaggerated body movements and expressions or confliction to common sense, which requires a higher-level understanding of human patterns and comedy-domain features. The recurrent neural network model could accomplish the task theoretically.

Still, the concept of humor is very abstract and subjective. Utilizing only the video information for detection may encounter the problems like data ambiguity and bad understanding of features.

*B. Difficulties in humor understanding and detection*

Modeling the humor via a unidimensional dataset could be very challenging. The key factors could be concluded as following:

- Most of the humor elements in videos are very creative. The original and fresh humor ideas could lead audience to laugh, while the old ones that basically everyone knows could barely receive laugher from audience. The creative actions is too ambiguous for neural networks to learn and recognize.

- The humor effects are always generated with the help of contextual information and environments. A humor punchline often needs progressive warm-ups before a dramatic performance or contradiction. The contextual information could be very important to identify humor punchlines. Thus the text data should be included into the data structure.

- Sometimes the humor may only detectable through audio effects. The different intonation of the same words could express various meanings and then lead to humor effects. The audio data also worth to be included in the data structure.

*C. Multimodal video understanding models*

The multimodal video understanding model is a fast growing field in natural language processing related tasks. In the three dimensions in this task, the text information possess both abstraction of the video meanings and the contextual information. Since the BERT model was introduced in 2018, the transformer-based models show state-of-the-art performances in many subtasks in natural language processing fields like question answering and natural language inferencing.

The multimodal model for humor detection in videos should focus more the textual understanding and feature detection.

*D. Previous research works*

In the newly proposed video understanding models, the BERT based models are considered to provide augmentation for attributes like timestamps [1]. Another existing approach that utilizing a cross class structure to better perceive the temporal concept receive the 1st place of the 3rd YouTube-8M Kaggle competition [2].

As for the textual humor detection research works, the transformer based models provide incredible performances [3], especially the Colbert model give out a 98.2% accuracy on short texts [4].

The cross-modal approach was introduced in 2017 as an effective textual and visual retrieval technique [5], which indicates that the strong cross modal information connection could be utilized for further implementations. Another approach integrates the audio and visual cures for computerized understanding is proposed to prove the possibility and capability of multimodal methods [6].

Among the previous multimodal research works, the Memory Fusion Network (MFN) model is propose in 2018 to conduct the multi-view sequential learning tasks [7]. In MFN, LSTM model is used to learn features separately in unimodal subtask. Then, they used a special attention mechanism named DMAN to learn interactive feature. At last, the Multi-view Gated Memory section summarized the temporal outcomes. The advantages of LSTM model are the contextual understanding of the texts and the maintenance of key features for long texts, which could provide a better understanding of specific task requirements. However, the MFN model did not take the shared contextual information among multimodal subtasks into consideration.

An improved model called Contextual Memory Fusion Network (C-MFN) proposed in 2019 fixed the neglected contextual information among multimodal subtasks using self-attention mechanism [8]. The paper tested the model performance on their new generated dataset called UR-FUNNY (it is the first humor detection dataset and its human performance is 82.5%). Their result outperformed basic MFN model (64.47%). And the best detection result is from the situation that train and test on all three modalities, which is 65.23%. The improvement is trivial and there is a big gap for models achieving the human-level performance.

In 2020, an new and effective approach called Multimodal Adaptation Gate (MAG) utilizing the pre-trained models from transformers like BERT and XLNet [9]. The MAG approach enable the pre-trained models to percept non-text data in the fine-tuning process. Its main idea is to pass a shift generated by visual and acoustic information to the pre-trained models. The experiments are conducted in the CMU-MOSI and CMU-MOSEI dataset designed for the multimodal sentiment analysis tasks, which achieves human-level performance.

*E. Pre-trained transformer models*

The BERT models are directly designed for the language understanding field through pre-training of deep bidirectional encoder representation methods. The token, segment and position embeddings in the BERT model could provide a contextual representation and understanding of the texts. The official site also provides multiple pre-trained BERT models to apply in different situations with different layers and pre-train data. The fine-tuning process is widely used in BERT pre-trained models to obtain the best performance on specific tasks.

However, the BERT models are not initially built to satisfy the multimodal tasks, which means that the necessary modifications is another key point for the model to accept and process two extra modalities.

The BERT also have its drawbacks. For one part, the Mask function in pre-train process would cause information loss when conducting fine-tuning steps. More specifically, the Mask function improves the generalization capability by masking features from useless embeddings but may thus cause the neglection of the same features being neglected in the fine-tuning steps and cause performance loss in specific domain tasks. For another part, the BERT assume that the masked words do not possess the contextual information, which could lead to general performance loss. Consequently, we also apply the XLNet model to avoid the possible performance loss.

Compared with the BERT model, the XLNet model applies the Permutation Language Model method to avoid the performance loss during pre-training process. The implementation of Transformer-XL features like relative location encoding and paragraph RNN improves the model performance in long-text tasks. The extended data amount and improved data quality for pre-training also contributes to the overall performance improvements. The XLNet model performs better than BERT in the classical NLP tasks like reading comprehension, text classification and information retrieval.

### III. DATASET SELECTION

#### A. Multimodal datasets with three modalities

- UR-FUNNY-1. The dataset is generated from the TED talks with various kinds of speakers and topics related to humor and it contains 1866 videos from 1741 different speakers and 417 topics [8]. There are 8257 positive samples with 8257 negative samples.

- The CMU-MOSI (Multimodal Opinion-level Sentiment Intensity). The dataset selected data from 93 videos randomly from 89 speakers in English. The 2199 videos clips are about 150 mins in total[10].

- The CMU-MOSEI (Multimodal Opinion Sentiment and Emotion Intensity). It is the improved version of CMU-MOSI dataset containing 1000 distinct speakers, 3228 videos, 250 different topics, 23453 video clips that are nearly 4000 mins in total.[11].

#### B. UR-FUNNY dataset

The UR-FUNNY dataset is accessible through ROC-HCI website (https://roc-hci.com/). The source of this dataset comes from the TED talks, which enables enough topic span and enriched meanings in the textual modality. The textual and acoustic information come from manual markers, which ensures the reliability of the data quality. There are also marks of audience behavior like laughter in the textual modality.

In addition, the dataset introduces concepts named punchline and context generated according to the laughter marker. The exact sentence before the marker is considered as the punchline and after the marker is the context.

The duration of the dataset is about 90 hours containing 1741 speakers and 417 distinct topics. The length of over 90% punchlines are less than 32. The following table shows the data distribution of the dataset. The percentage is well chosen to get the best performance during fine tuning.

TABLE I. STATISTICS OF UR-FUNNY DATASET

| Statistics | Data distribution | | |
|---|---|---|---|
| | *Train* | *Val* | *Test* |
| Number of Humor instance | 5306 | 1313 | 1638 |
| Number of none-Humor instance | 5292 | 1313 | 1652 |

#### C. CMU-MOSI dataset

The videos are collected from the YouTube channels that expression opinions about various topics. The collected videos basically only contains only one human face looking the camera in each of the videos, which largely avoids the biases caused by unrelated objects during the fine tuning process of the model.

The data in textual modality came from the manually transcribed video contents, including the pause words like umm. The data in acoustic modality was generated using P2FA tool (https://github.com/ucbvislab/p2fa-vislab). The overall data generation process was checked using PRAAT at the final stage [12]. The annotated visual features mainly focus on the human face expressions.

The subjective opinions was labelled in the conditions like directly mentioning, referencing and implicitly indicating. The instances in the dataset were further segmented if it contains multiple entities, sentiment adjustments or ambiguities. The dataset contains only the subjective sentiment instances with on objective ones. The sentiment intensity of 2199 short subjective clips were defined from -3 to +3, which means strongly negative to strongly positive. Annotation work was done by master human workers.

#### D. UR-FUNNY dataset

This dataset is basically similar to the CMU-MOSI dataset but using a larger scale of data and more advanced techniques to extract features in modalities. The large amount of the videos was collected from YouTube and the collections of each channel is under 10 clips with clear transcripts, which ensures the diversity of the data. The textual data used punctuation markers rather than the Stanford CoreNLP tokenizer for the better sentence quality [13].

Compared with CMU-MOSI dataset, CMU-MOSEI dataset introduced the emotion dimension {happiness, sadness, anger, fear, disgust, surprise} with 0 to 3 annotation scale.

### IV. DATASET ANALYSIS AND EVALUATION

#### A. UR-FUNNY dataset

The manual generated labels in this dataset provides an outstanding quality for researcher to check whether their developed models could perform human level understanding of the humor in multimodal tasks (accuracy of 82.5%).

The introduced data structure of punchline and context do contribute to let the LSTM based models better perceiving the key information and features exactly around the manually labelled markers. However, the LSTM models may not obtain the general context features but just the features and patterns around markers, which could lead to some overfitting issue during training. The LSTM based models may perform well in this dataset, but the generalization ability may not good enough in the large-scale implementation scenarios. Besides, the data structure of UR-FUNNY cannot directly fit in the basic pre-trained bidirectional transformers models.

In this humor detection task, it is not appropriate to develop the pre-trained bidirectional transformers models in the UR-FUNNY dataset unless make extra structure level modifications that leave only the labels with data from three modalities.

## B. CMU-MOSI dataset

Compared with the UR-FUNNY dataset, the MOSI dataset owns a simpler data structure for the models to conduct fine tuning process. The subjective sentiment instances could then evaluate the model's performance strictly on the human abstract emotion understanding and detection scenarios.

However, the insufficiency of facial gestures types (only 4) and total video numbers (only 93) means that the dimension and the quantity of instances is not big enough for the models to learn the patterns between three modalities.

## C. CMU-MOSEI dataset

Compared with the CMU-MOSI dataset, CMU-MOSEI dataset firstly introduced the emotion dimension to enable the model to understand not only from the sentiment perspective but also the emotion perspective of the human behaviors. This could be a great step for models understanding the complex and abstract human patterns and then precepting human humor behaviors.

## D. Experiment design

It could be a good test to evaluate whether the developed model possesses the ability to understand the human multimodal sentiments using CMU-MOSI dataset. After that, the CMU-MOSEI dataset can help to evaluate the model's capability of understanding in-depth and abstract human patterns using sentiment and emotion indicators.

## V. MODEL DESIGN

### A. Multimodal Adaptation Gate (MAG) Mechanism

In the multimodal understanding models, the key point is to enable the model to perceive the data from all modalities simultaneously.

The attention over all three modalities is proposed to obtain the fused features in the data as the first layer of input. After the fusion process, the attention gate adds the features in to the lexical inputs as a shift. The lexical features are the core input for model.

For the model do not have the none lexical shift, understanding of the word is completely determined by the word semantics and contextual lexical information. However, with the help of the added none lexical shift, the model could perceive the visual and acoustic features alongside the pure text. The adaptation would then help the model to obtain a fused view of the original text features not using shift mechanism.

The MAG mechanism is one simple feature adjustment function applied in one layer during the fine-tuning process of BERT model, which can work in any layer of the whole model.

### B. MAG-BERT

The pre-trained models and MAG fusion function form the overall structure of the model. The BERT has lexical vectors in each layer. The amendment shift caused by the MAG function allows the word vectors inside the BERT layers to take up the none lexical features during fine-tuning process.

### C. MAG-XLNet

Similar to the MAG-BERT model, the MAG-XLNet model simply use the different pre-trained model. The XLNet embedders and encoder could give different performance on the whole task. In the model level mechanism, the XLNet model could perform better in long text understanding theoretically.

### D. Wandb

In the experiment process, we choose the Wandb framework to visualize the performance of the models in charts rather than the TensorBoard framework [14].

## VI. IMPLEMENTATION AND PERFORMANCE

### A. Implementation

The implementation steps are visible and straight forward in the Jupyter notebook files. The Data-check notebook provide the data structure check and validation steps with executable codes. The Implementation notebook provides the environment and software configuration steps with codes and the training commands with outputs.

Implementation environment: Python 3 / Pytorch / RAM 26 GB / Disk storage 147 GB / GPU Tesla P100 16GB

Implementation settings: dropout 0.5 / fine_tuning_epoch 40 / learning rate 1e-5

### B. Result of experiments

TABLE II.    EXPERIMENT RESULT FOR CMU-MOSI DATASET

| Model | Performance | | | | |
|---|---|---|---|---|---|
| | *Accuracy* | *F1_score* | *MAE* | *Corr* | *Runtime* |
| MAG-BERT - mosi | 0.83664 | 0.83683 | 0.7636 | 0.7801 | **217s** |
| MAG-XLNet - mosi | **0.84885** | **0.84927** | **0.7634** | **0.8118** | 1497s |

TABLE III.    EXPERIMENT RESULT FOR CMU-MOSEI DATASET

| Model | Performance | | | | |
|---|---|---|---|---|---|
| | *Accuracy* | *F1_score* | *MAE* | *Corr* | *Runtime* |
| MAG-BERT - mosei | 0.82762 | 0.82822 | **0.6291** | 0.7636 | **4051s** |
| MAG-XLNet - mosei | **0.84227** | **0.84321** | 0.6417 | **0.7798** | 5340s |

The accuracy and the f1 score are straight forward to understand, higher means better. The MAE is the mean-absolute error, lower means better. The Corr is the Pearson Correlation, higher means better. The best parameter is bolden in the table. The detailed plots generated during fine tuning process are attach to the models using hyperlinks in the tables.

**[Personal experience and thoughts**

*According to the paper that I reproduce, the MAG-XLNet does perform better (exceeds at least 3% accuracy) than other baseline models like Tensor Fusion Network, Multi-attention Recurrent Network, Memory Fusion Network, Recurrent Memory Fusion Network and Multimodal Transformer for unaligned Multimodal Language Sequence [3]. The outcome is in accordance with my previous theoretical analysis.*

*However, I only reproduced the MAG-BERT and MAG-XLNet models, so I did not directly use experiment data from others.*

**Personal experience and thoughts]**

## VII. PERFORMANCE EVALUATION

### A. Datasets perspective

*1) CMU-MOSI dataset*

The CMU-MOSI dataset is a small dataset for confirming the designed models possess the capability of understanding human's sentiments with the help of MAG shift function and pre-trained transformer models. The XLNet model performs better than BERT model. But the only 1% performance improvement from XLNet model requires much more runtime. In the actual application scenarios, the time consumption links directly to the resource costs, which is a very important aspect for business applications.

In the small dataset (less than 5000 clips), the MAG-BERT model would be a better choice when considering the time consumption issue.

*2) CMU-MOSEI dataset*

The instance amount of CMU-MOSEI dataset is 10 times larger than the CMU-MOSI dataset. When fine tuning on the large dataset, the advantages of XLNet model shows up. It achieves a performance improvement at about 2% and consumes nearly the same run time compared with BERT model.

### B. Models perspective

The MAG-XLNet model performs best in the multimodal emotion detection tasks but would basically cost more time consumption which could lead to cost increasement in actual application scenarios. However, when handling a small set of data, the MAG-BERT could be a better choice for its similar outstanding performance using much less time.

## VIII. FUTURE IMPROVEMENTS

### A. Dataset improvements

The CMU-MOSI and CMU-MOSEI dataset are for general multimodal sentiment and emotion analysis. They are sufficient to test the capabilities of models on abstract human behavior understanding. It is good to find out that the basic MAG mechanism could perform well with the pre-trained models. Still, more specific tests in different datasets could help to locate more valuable spots in humor detection fields.

### B. Technical improvements

Theoretically, the MAG mechanism could be applied in any layer in the model. The lexical shift components can exclude the text features to make the model focus less on the text and more on audio and video features. The fusion function can also be placed for several times but not only one time.

Besides, other technical mechanism can also be designed and applied in the multimodal sentiment analysis and humor detection tasks.

## IX. CONCLUSION

In this paper, we firstly analyzed the benefits of studying the abstract and in-depth video understanding tasks. A data structure that includes the information of video frames, audio spectrum and texts provide a new direction to explore. We then elaborated the current research progresses in multimodal sentiment analysis on text, video and audio modalities. The feature detection methods for abstract category remains blank. After that, we mentioned the drawbacks of traditional video understanding methods and the difficulties of developing the in-depth video emotion detection models. In the next part of our paper, we proposed a Multimodal Adaptation Gate mechanism to fuse the data in three modalities into one form. We choose the transformer based pre-trained models to apply the MAG function because those models have already learnt how to understand the general textual information with contextual embeddings. At last, we tested and evaluated the MAG-BERT and MAG-XLNet models' performance in CMU-MOSI and CMU-MOSEI dataset. We also concluded some suggestions for implementations in the actual scenarios and the possible improvement directions in the future research works.

## ACKNOWLEDGMENT

Thanks to Professor Pan's informative and inspiring courses and his kindly instruction and help during my completing the research works in this interesting subject. Thanks to my supporting family that provide the financial help to buy the experiment equipment. I am also very grateful to have my nice and warm-hearted friends and classmates who gives me many mental supports and encouragements.